\documentclass{article}

\usepackage[utf8]{inputenc}
\usepackage[parfill]{parskip}
\usepackage[T1]{fontenc}
\usepackage[tt=false,type1=true]{libertine}
\usepackage[varqu]{zi4}
\usepackage{amsmath,amssymb,amsthm,mathtools}
\usepackage[libertine]{newtxmath}
\usepackage{microtype}

\usepackage{amsmath,amsfonts,bm}

\def\eqref#1{equation~\ref{#1}}

\def\1{\bm{1}}

\DeclareMathAlphabet{\mathsfit}{\encodingdefault}{\sfdefault}{m}{sl}
\SetMathAlphabet{\mathsfit}{bold}{\encodingdefault}{\sfdefault}{bx}{n}

\usepackage{graphicx}
\usepackage[font=small]{caption}
\usepackage{subcaption}
\usepackage{float,afterpage,newfloat,placeins,wrapfig}
\usepackage{adjustbox}
\usepackage{booktabs,multirow,makecell,array,tabularx,threeparttable}
\usepackage{xcolor,colortbl}

\usepackage{enumitem}
\usepackage{listings}
\usepackage{comment}
\usepackage{xspace}
\usepackage{minitoc}

\usepackage[disable]{todonotes}

\usepackage{authblk}

\usepackage{natbib}
\setcitestyle{authoryear,round,citesep={;},aysep={,},yysep={;}}

\usepackage{url}
\usepackage{hyperref}
\definecolor{darkblue}{rgb}{0,0,0.5}
\hypersetup{
    colorlinks=true,
    citecolor=darkblue,
    linkcolor=darkblue,
    urlcolor=darkblue,
    pdftitle={ConvCue: Complementary Visual Inductive Biases for Vision-Language Models},
    pdfauthor={Zixuan Lan and Shichu Sun}
}

\newcommand{\name}{\textsc{ConvCue}\xspace}

\lstdefinestyle{pythonstyle}{
    language=Python,
    basicstyle=\ttfamily\footnotesize,
    breaklines=true,
    showstringspaces=false
}

\title{ConvCue: Complementary Visual Inductive Biases for Vision-Language Models}

\author[1]{Zixuan Lan}
\author[2]{Shichu Sun}
\affil[1]{The University of Chicago}
\affil[2]{University of Chinese Academy of Sciences}
\date{}

\begin{document}
\pagestyle{plain}
\maketitle

\begin{abstract}
\noindent
Modern vision-language models (VLMs) achieve strong performance across a broad range of multimodal tasks, yet still struggle with visual questions that require fine-grained discrimination and spatial understanding. These limitations motivate investigating whether supplementary visual representations can improve existing VLMs without replacing their native visual encoders. Pretrained convolutional networks offer a candidate feature source, motivated by their local connectivity and spatial weight sharing. We introduce CONVCUE, which augments the native visual representations of a pretrained VLM with final-stage features from a parallel, frozen pretrained CNN. A learnable adapter maps convolutional features to the native visual feature dimension, while gated cross-attention allows the original visual tokens to retrieve information from the CNN features. The enhanced tokens are passed through the original visual-to-language projector, and the model is adapted through a two-stage training procedure. We evaluate CONVCUE on Qwen3-VL-2B, Qwen3-VL-4B, and LLaVA-OneVision-7B across 13 multimodal benchmarks covering visual question answering, document and chart understanding, and multimodal reasoning. CONVCUE improves average benchmark performance over both the original models and matched two-stage fine-tuning controls on all three backbones. On Qwen3-VL-4B, it improves over the original model on all 13 benchmarks and raises the average score from 75.00 to 78.82 relative to the matched fine-tuning control. These results show that pretrained convolutional representations, when integrated through learned adaptation and fusion, can improve the visual understanding of existing VLMs without replacing their original visual encoders. Code will be released at \url{https://github.com/LanLanLan-Ian/CONVCUE}.
\end{abstract}

\section{Introduction}

Vision-language models (VLMs) have made great progress across a broad range of tasks, yet they still make mistakes on simple visual questions. Strong performance on multimodal tasks does not always mean that a model can reliably distinguish what it sees. Prior work shows that visual encoders can produce similar representations for visually distinct images, and that downstream VLMs can struggle with questions that depend on these distinctions \citep{tong2024eyes, lan2026seeing}. These findings raise a practical question: can integrating supplementary visual representations into pretrained VLMs improve their downstream visual understanding?

A visual representation may support its training task without making all the distinctions needed to answer downstream questions. For example, consider two images of the same building, one with an intact window and the other with broken glass. Both images may match the caption \textbf{a building with glass windows}, but answering \textbf{Is the window intact?} requires distinguishing local details such as cracks and missing glass, rather than merely recognizing the building and its windows. Prior work shows that models can score well on image--text retrieval yet struggle to distinguish captions that differ in object attributes, relationships, or word order \citep{yuksekgonul2023visionlanguagemodelsbehavelike}. Thus, strong retrieval performance alone does not guarantee sensitivity to the specific attributes and relationships queried in downstream tasks. The visual information expressed in these representations depends on how the encoder is trained and how it aggregates image features. For example, DINO finds that object regions appear more clearly in the attention maps of self-supervised ViTs than in those of the supervised ViTs examined \citep{caron2021emergingpropertiesselfsupervisedvision}. Studies comparing ViTs and ResNets also find differences in how local and global information is represented: ViTs incorporate global information in early layers, while ResNets aggregate local features over progressively larger regions \citep{raghu2022visiontransformerslikeconvolutional}. These observations motivate investigating whether representations from a pretrained CNN can provide useful supplementary information for an existing VLM.

\begin{figure*}[t]
    \centering
    \includegraphics[width=\textwidth]{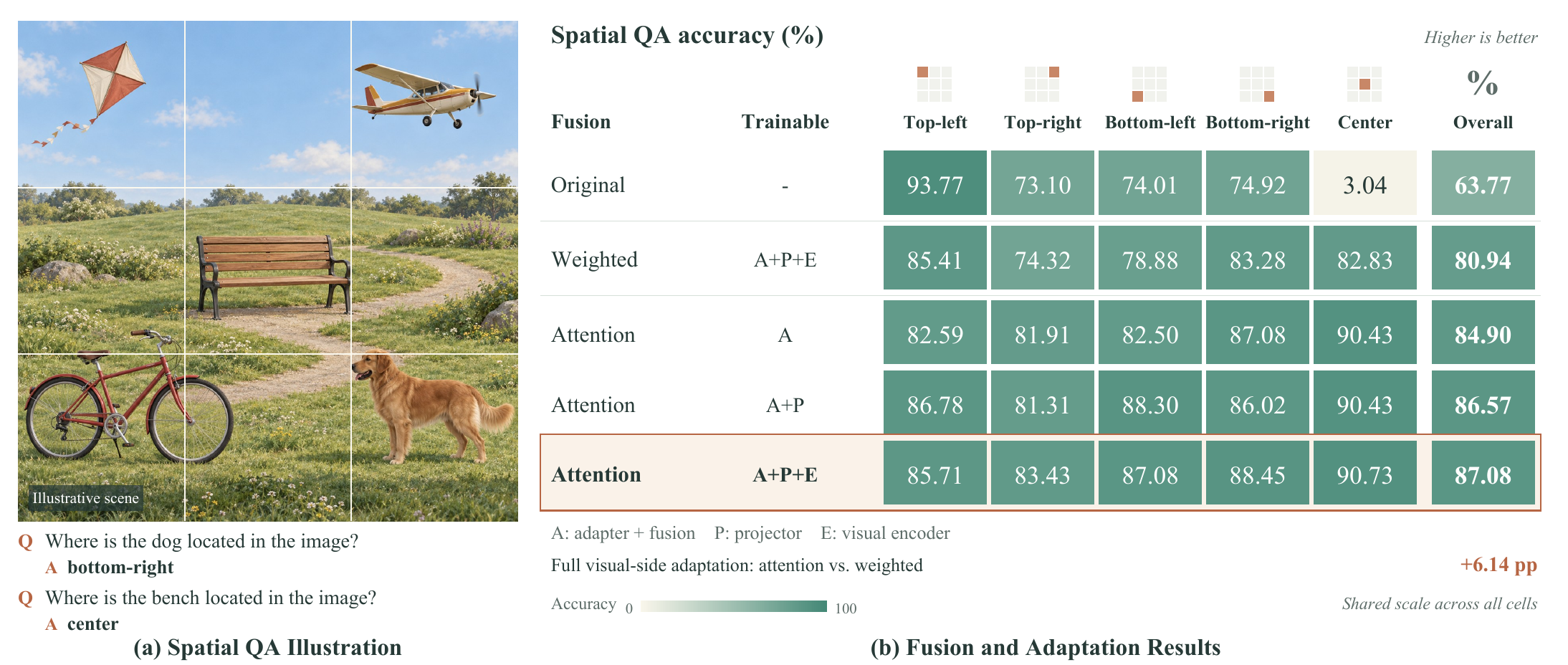}
    \caption{
    Pilot experiments on Spatial QA.
    (a) Illustration of the example question--answer pairs.
    (b) Overall and per-position accuracy for different fusion strategies and trainable component configurations. A denotes the adapter and fusion module, P the original projector, and E the original visual encoder. The outlined row highlights the full visual-side adaptation setting, under which cross-attention outperforms weighted fusion by 6.14 percentage points.
    }
    \label{fig:pilot_spatial_qa}
\end{figure*}

Combining different visual representations is an established direction in multimodal modeling. Eyes Wide Shut combines CLIP features with features learned through visual self-supervision, exploring how representations learned with different objectives can support visual understanding together \citep{tong2024eyes}. Cambrian-1 studies multiple visual encoders and introduces a spatially aware module to aggregate their features for language models \citep{tong2024cambrian}. Eagle examines which encoders to combine, how to integrate their features, and how to align them with the language model before joint training \citep{shi2025eagleexploringdesignspace}. LLaVA-HR combines low-resolution ViT and high-resolution CNN pathways through mixture-of-resolution adapters \citep{luo2025feast}. Together, these studies show that improving visual representations is not limited to building stronger individual encoders: features from different sources can also be brought together to support multimodal understanding. This direction involves choosing supplementary features, designing how they interact with the original representations, and adapting the combined features for use by the existing model.

Following this direction, we investigate pretrained convolutional representations as a supplementary feature source for the visual pathways of existing VLMs. Convolutional networks explicitly incorporate two-dimensional neighborhood structure into feature extraction through local connectivity and spatial weight sharing, providing architectural motivation for this exploration \citep{dosovitskiy2020image,lahoti2024rolelocalityweightsharing}. To this end, we introduce \name{}, a visual representation augmentation framework that integrates convolutional feature extraction, representation adaptation, and gated fusion. \name{} retains the original visual encoding pathway and processes the same source image with a separate frozen pretrained CNN branch. A learnable adapter maps the CNN's final-stage features to the native visual feature dimension, after which the original visual tokens aggregate supplementary information through cross-attention and incorporate it through gated residual updates. The enhanced visual representations are passed to the language model through the original visual-to-language projection module, allowing the supplementary branch to contribute to downstream predictions without replacing the original visual encoder.

We evaluate \name{} on Qwen3-VL-2B, Qwen3-VL-4B, and LLaVA-OneVision-7B across 13 visual understanding benchmarks. Across all three backbones, \name{} achieves higher average performance than both the original models and matched Direct SFT controls, demonstrating gains from the integrated augmentation design beyond those obtained through the same additional training. On our main Qwen3-VL-4B backbone, \name{} improves over the original model on all 13 benchmarks and also outperforms Qwen3-VL-8B on each benchmark. Pilot experiments examine fusion strategies and the scope of visual-side adaptation, while subsequent representation analyses and a case study examine changes in visual features and answer preferences. Together, these results show that pretrained convolutional representations, when integrated through learned adaptation and fusion, can improve the visual understanding of existing VLMs without replacing their original visual encoders.

\section{Pilot Experiments}
\label{sec:pilot_experiments}

Adding convolutional representations to an existing VLM requires choosing how to fuse the features and which components to train. We first explore these two design choices through small-scale experiments to guide the subsequent framework. We use LLaVA-1.5-7B \citep{liu2023visual} and Spatial QA constructed from MS COCO \citep{lin2015microsoftcococommonobjects}, keeping the language model and CNN frozen. Figure~\ref{fig:pilot_spatial_qa} presents the task illustration and results. Dataset construction is detailed in Appendix~\ref{app:spatial_qa_construction}.

\noindent\textbf{An initial comparison of fusion strategies.}
We initially use weighted fusion to directly combine the adapted convolutional features with the original visual features after aligning their spatial grids. As an alternative, we consider cross-attention, which uses the original visual tokens as queries to aggregate convolutional features without requiring one-to-one spatial correspondence. This comparison evaluates the two fusion configurations, including their respective spatial alignment procedures. In both configurations, we train the original visual encoder, projector, and added learnable components. Cross-attention achieves an overall accuracy of 87.08\%, compared with 80.94\% for weighted fusion, and obtains higher accuracy in all five position categories. These results provide initial support for adopting the cross-attention configuration in our subsequent framework.

\noindent\textbf{Exploring the scope of visual-side adaptation.}
For the cross-attention configuration, we first train only the added adapter and fusion module, keeping the original VLM parameters fixed. The fused features still pass through the original projector, while the visual tokens used for fusion come from the original visual encoder. We therefore also examine configurations that allow these components to adapt. Training only the added modules yields an overall accuracy of 84.90\%. Including the projector raises accuracy to 86.57\%, and additionally training the visual encoder raises it to 87.08\%. Among the configurations examined, allowing the original visual components to adapt yields higher overall accuracy, with a smaller improvement from additionally training the visual encoder. This observation motivates jointly adapting the added modules and the original visual pathway in the subsequent framework. These pilot results guide our initial design choices; they do not establish their effects on other models or tasks. We incorporate the selected fusion strategy and visual-side adaptation configuration into \name{} and evaluate the complete framework in broader experiments.

\section{Method}
\subsection{Preliminaries}
\label{sec:preliminaries}

\begin{figure*}[t]
    \centering
    \includegraphics[width=\textwidth]{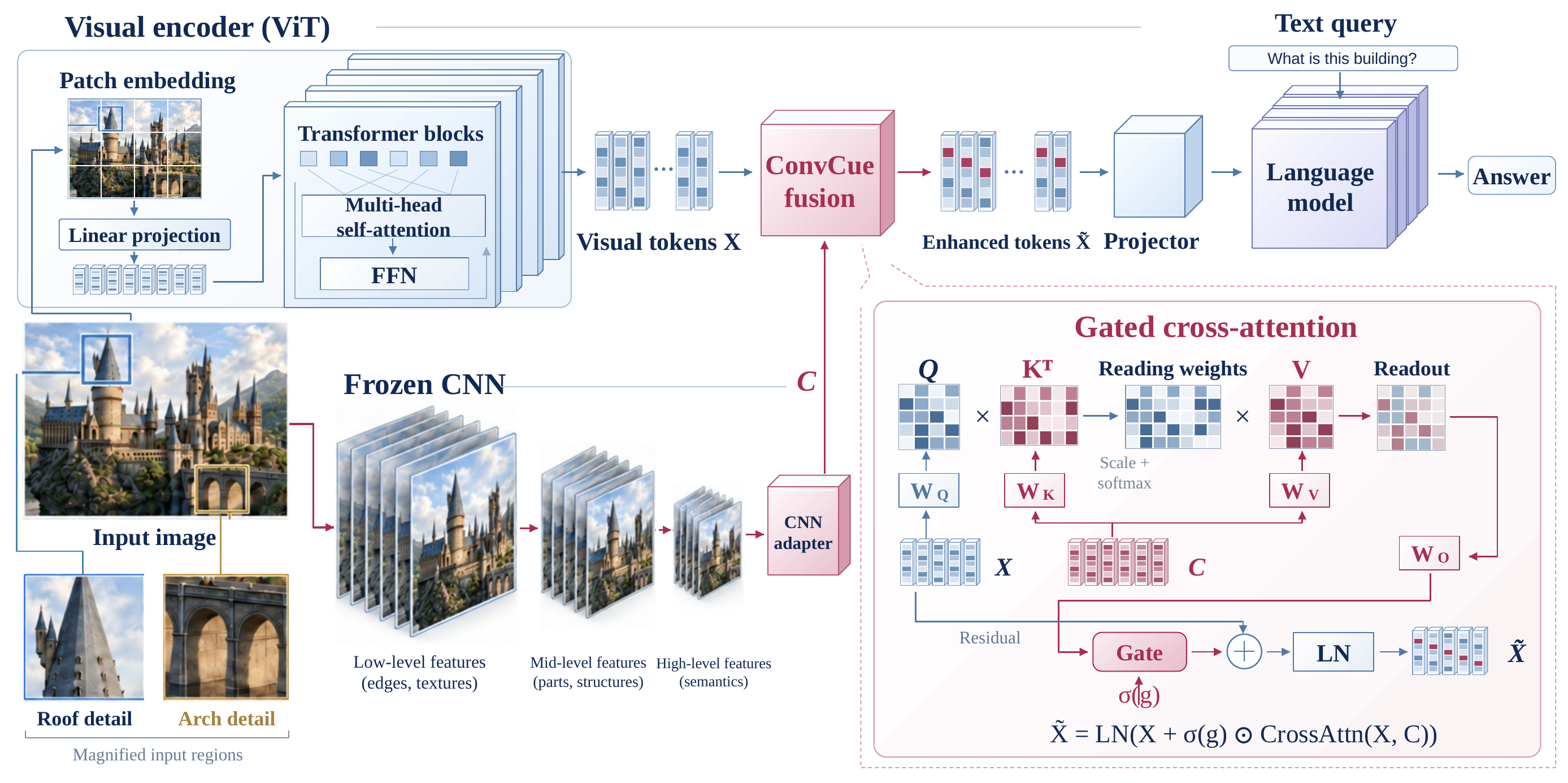}
    \caption{Overview of \name{}. The original visual encoder processes image patches, while a frozen CNN extracts convolutional features from the same image and a trainable adapter maps them to the visual token dimension. Gated cross-attention uses the original visual tokens as queries and the adapted CNN features as keys and values. A channel-wise gate controls the residual update, followed by LayerNorm. The enhanced tokens then enter the original visual-to-language mapping and language model. The enlarged panel details the fusion mechanism; feature grids and token colors are schematic.}
    \label{fig:method_overview}
\end{figure*}

We consider a vision-language model composed of a visual encoder, a visual-to-language projector, and a language model \citep{liu2023visual}.
The model takes an image $I$ and a text instruction $q$ as input and generates a textual response $Y$. The visual encoder $E_v$ first processes the image to produce visual features $X = E_v(I) \in \mathbb{R}^{N \times d_v}$,  where $N$ is the number of visual tokens and $d_v$ is their feature dimension. Each token corresponds to a spatial location in the image and may also encode contextual information acquired through the visual encoder. The projector $P$ maps these features into representations compatible with the language model, $Z = P(X) \in \mathbb{R}^{M \times d_\ell}$, where $d_\ell$ is the language model's embedding dimension and $M$ is the number of projected visual tokens.
This mapping may include spatial token aggregation, so $M$ need not equal $N$.
Together, $Z$ and the token embeddings of $q$ form the multimodal input, conditioned on which the language model autoregressively generates $Y$.
Throughout this paper, \emph{visual tokens} refer to the features $X$ before projection. Building on the pilot observations, we introduce \name{}, which augments these tokens using a pretrained CNN branch, a learnable adapter, and gated cross-attention. The enhanced representations are passed to the original projector, preserving the model's existing visual-to-language pathway. Figure~\ref{fig:method_overview} illustrates the framework.

\subsection{Convolutional Feature Extraction and Adaptation}
\label{sec:cnn_adaptation}

To complement the original visual tokens $X$, a separate pretrained CNN encoder $E_c$ processes the same image $I$ and extracts its final-stage feature map $F_c = E_c(I) \in \mathbb{R}^{d_c \times h \times w}$, where $d_c$ is the channel dimension and $h$ and $w$ denote the spatial dimensions.
A learnable CNN adapter maps these features to the visual token dimension $d_v$ while preserving their spatial grid. Specifically, a $1\times1$ convolution projects the channels from $d_c$ to $d_v$, followed by spatial flattening and LayerNorm to obtain $U = \operatorname{LN}_1(\operatorname{Flatten}(\operatorname{Conv}_{1\times1}(F_c))) \in \mathbb{R}^{N_c \times d_v}$, where $N_c = hw$.
The adapted tokens are then computed as $C = \operatorname{LN}_2(U + \operatorname{MLP}(U))$, using a two-layer MLP with a GELU activation and hidden dimension $4d_v$.
The output $C \in \mathbb{R}^{N_c \times d_v}$ serves as the feature sequence for subsequent cross-attention fusion, without requiring $N_c$ to match the original visual token count $N$.
The pretrained CNN remains frozen throughout training, while the adapter is trained in both stages.
\subsection{Gated Cross-Attention Fusion}
\label{sec:gated_fusion}

Given the original visual tokens $X \in \mathbb{R}^{N \times d_v}$ and the adapted CNN tokens $C \in \mathbb{R}^{N_c \times d_v}$, we compute $A = \operatorname{MHA}(X, C, C) \in \mathbb{R}^{N \times d_v}$, using $X$ as queries and $C$ as keys and values.
Here, $\operatorname{MHA}$ includes the query, key, and value projections for each attention head and the output projection.
Each visual token involved in fusion can attend to all CNN tokens based on their feature content, without requiring equal token counts or a one-to-one spatial correspondence. To control the convolutional update, we introduce a learnable channel-wise gate $g = \sigma(a)$, where $a \in \mathbb{R}^{d_v}$ and $\sigma$ denotes the sigmoid function.
The gate is shared across token positions and images, learning a separate scaling factor for each channel rather than being dynamically generated for each input.
The fused representation is computed as $\widetilde{X} = \operatorname{LN}(X + g \odot A)$, where $\odot$ denotes element-wise multiplication with $g$ broadcast across token positions.
This gated residual update preserves the shape of $X$, and the enhanced tokens are passed through the original projector to obtain $Z = P(\widetilde{X})$.

\subsection{Training}
\label{sec:training}

We train \name{} in two stages: vision-language alignment and joint supervised fine-tuning.
The pretrained CNN encoder remains frozen throughout both stages. \textbf{Vision-language alignment.}
We freeze the language model and jointly train the original visual encoder, CNN adapter, gated cross-attention fusion module, and projector using image--text pairs.
This stage adapts the convolutional features to the original visual representations and aligns the fused features with the frozen language model. \textbf{Joint supervised fine-tuning.}
We initialize from the alignment checkpoint and additionally unfreeze the language model.
Using visual instruction data, we jointly update the visual encoder, adapter, fusion module, projector, and language model, allowing visual feature extraction, fusion, and language generation to adapt together. Both stages optimize the autoregressive objective $\mathcal{L} = -\frac{1}{T}\sum_{t=1}^{T}\log p_{\theta}(y_t \mid I, q, y_{<t})$, where $Y=(y_1,\ldots,y_T)$ is the target response and $\theta$ denotes the parameters optimized in the current stage.
Training uses ground-truth preceding response tokens, and the loss is computed only over target response tokens, excluding image and instruction positions.

\section{Experiments}
\subsection{Experimental Setup}
\label{sec:experimental_setup}

\begin{figure*}[t]
    \centering
    \includegraphics[width=\textwidth]{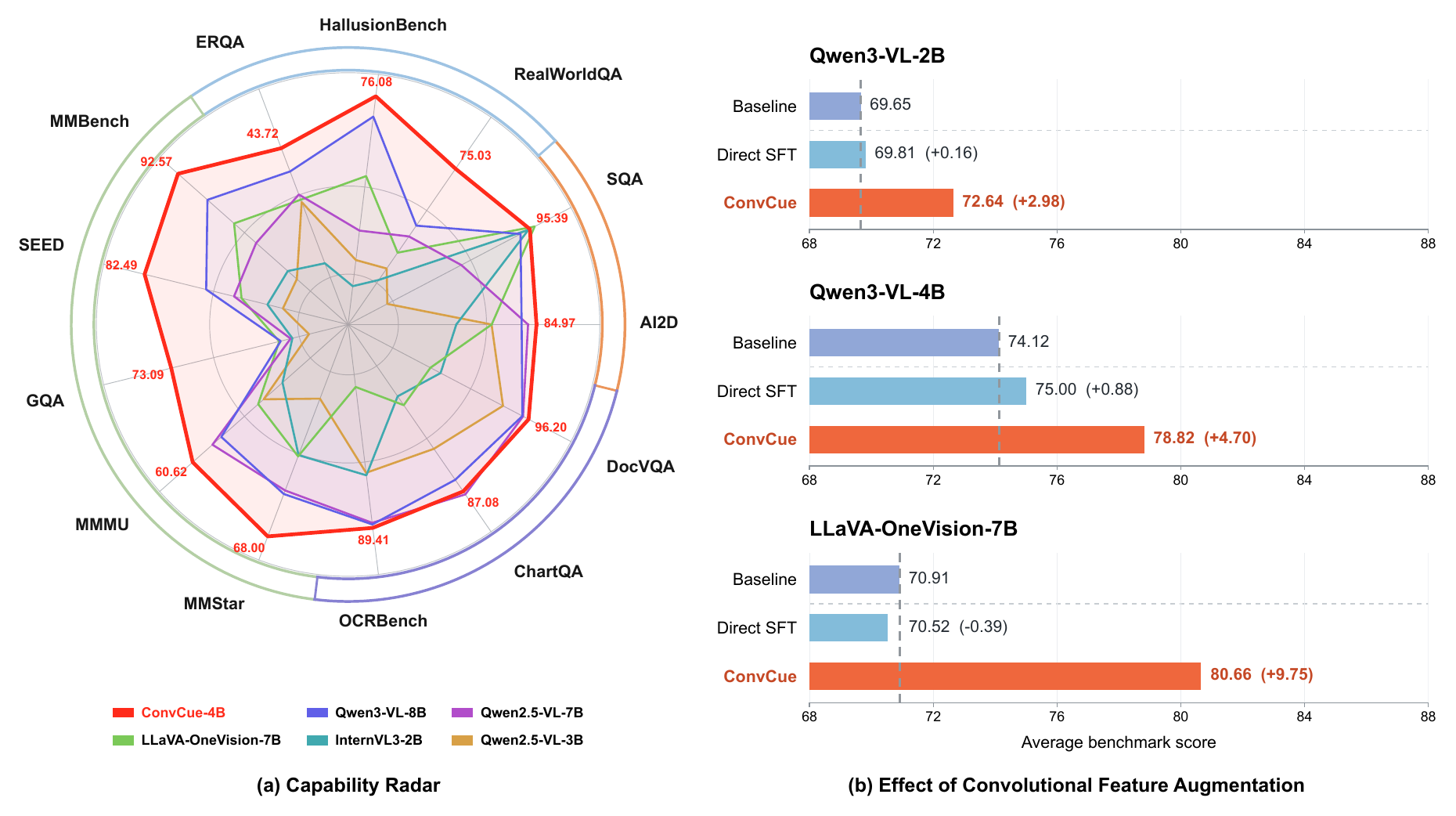}
    \caption{(a) Comparison across 13 visual understanding benchmarks. ConvCue uses Qwen3-VL-4B; red labels indicate its scores. (b) Average scores across the same 13 benchmarks for the original models, direct supervised fine-tuning, and ConvCue. Parentheses show score-point changes from each original model; dashed vertical lines mark its average score.}
    \label{fig:capability_ablation}
\end{figure*}

Our main model, \name{}-4B, is built on Qwen3-VL-4B \citep{bai2025qwen3}. We compare it across 13 benchmarks with GPT-4.1 and GPT-4V \citep{gpt41}, Claude-3.7-Sonnet \citep{claude37}, Gemini-1.5-Pro \citep{geminiteam2024gemini15unlockingmultimodal}, and open-source models including Qwen2.5-VL \citep{bai2025qwen25vltechnicalreport}, InternVL3/3.5 \citep{zhu2025internvl3exploringadvancedtraining,wang2025internvl35advancingopensourcemultimodal}, LLaVA-OneVision \citep{li2024llava}, Gemma3 \citep{gemmateam2025gemma3technicalreport}, DeepSeek-VL2 \citep{wu2024deepseekvl2mixtureofexpertsvisionlanguagemodels}, MiMo-VL \citep{coreteam2025mimovltechnicalreport}, Phi-4-Multimodal \citep{microsoft2025phi4minitechnicalreportcompact}, MiniCPM-V-4 \citep{yao2024minicpm}, Kimi-VL \citep{kimiteam2025kimivltechnicalreport}, SAIL-VL \citep{dong2025scalablevisionlanguagemodel}, and Ovis2 \citep{lu2024ovisstructuralembeddingalignment}. Ablations compare the original model, direct supervised fine-tuning, and \name{} on Qwen3-VL-2B \citep{bai2025qwen3}, Qwen3-VL-4B \citep{bai2025qwen3}, and LLaVA-OneVision-7B \citep{li2024llava}. We use the LLaVA image--text pretraining data for vision-language alignment in the first stage \citep{liu2023visual}, followed by the visual instruction data used in LLaVA-UHD for joint supervised fine-tuning in the second stage \citep{guo2024llava}. The training data are shared across the three model backbones. Data processing procedures and training hyperparameters are provided in the appendix. We evaluate on 13 visual understanding benchmarks: AI2D \citep{kembhavi2016diagram}, SEED-Bench \citep{li2023seed}, HallusionBench \citep{guan2024hallusionbench}, RealWorldQA \citep{xai2024grokvision}, ScienceQA \citep{lu2022learn}, OCRBench \citep{liu2024ocrbench}, ERQA \citep{team2025gemini}, MMBench \citep{liu2024mmbench}, MMMU \citep{yue2024mmmu}, ChartQA \citep{masry2022chartqa}, DocVQA \citep{mathew2021docvqa}, GQA \citep{hudson2019gqa}, MMStar \citep{chen2024we}. For models evaluated in our experiments, we use greedy decoding
without stochastic sampling. We additionally compare \name{} with Cambrian-1-8B \citep{tong2024cambrian}, LLaVA-UHD-v3-7B \citep{sun2025llavauhdv3progressivevisual}, Eagle-X4-8B-Plus \citep{shi2025eagleexploringdesignspace}, and Florence-VL-8B \citep{chen2024florencevlenhancingvisionlanguagemodels} on nine shared benchmarks. All the tests follow their paper.

\begin{table*}[t]
\centering
\caption{Comparison on visual-language benchmarks.}
\label{tab:vlm_comparison}
\setlength{\tabcolsep}{3pt}
\renewcommand{\arraystretch}{1.12}
\resizebox{\textwidth}{!}{%
\begin{tabular}{lcc*{13}{c}}
\toprule
Model & Scale & Avg. & MMBench & SEED & RealWorldQA & SQA & AI2D & HallusionBench & OCRBench & ERQA & MMMU & ChartQA & DocVQA & GQA & MMStar \\
\midrule
\multicolumn{16}{l}{\textit{Closed-source MLLMs}} \\
GPT-4.1-nano-20250414 & -- & 58.47 & 62.40 & -- & -- & 65.90 & 68.00 & 36.90 & 71.00 & -- & 57.60 & -- & -- & -- & 47.50 \\
GPT-4.1-mini-20250414 & -- & 67.68 & 80.90 & -- & -- & -- & 76.00 & 49.30 & 84.00 & -- & 55.00 & -- & -- & -- & 60.90 \\
GPT-4V (0409) & -- & 70.75 & 79.80 & 73.00 & 68.00 & 84.80 & 78.60 & 43.90 & 65.60 & -- & 61.70 & 78.50 & 88.40 & -- & 56.00 \\
Claude-3.7-Sonnet & -- & 72.18 & 79.70 & 74.30 & 55.40 & 90.90 & 82.50 & 55.40 & 70.10 & 35.50 & 71.00 & 92.20 & 94.10 & -- & 65.10 \\
Gemini-1.5-Pro & -- & 72.71 & 73.90 & 76.00 & 64.10 & 85.70 & 79.10 & 45.60 & 75.40 & -- & 60.60 & 87.20 & 93.10 & -- & 59.10 \\
\midrule
\multicolumn{16}{l}{\textit{Open-source MLLMs}} \\
Gemma-3-4B-IT & 4B & 58.36 & 66.40 & 65.39 & 44.00 & 67.20 & 70.70 & 40.80 & 66.00 & -- & 47.30 & 68.80 & 75.80 & 40.00 & 47.90 \\
InternVL3-1B & 1B & 63.09 & 68.20 & 71.16 & 58.20 & 89.84 & 69.70 & 37.20 & 79.80 & 30.30 & 43.20 & 75.30 & 81.90 & -- & 52.30 \\
InternVL3-2B & 2B & 68.20 & 78.00 & 74.95 & 64.30 & 95.23 & 78.60 & 41.90 & 83.10 & 31.50 & 48.70 & 80.20 & 88.30 & 60.70 & 61.10 \\
Qwen2.5-VL-3B & 3B & 68.37 & 76.80 & 74.00 & 65.40 & 79.40 & 81.40 & 46.60 & 82.80 & 38.00 & 51.20 & 84.00 & 93.90 & 59.00 & 56.30 \\
DeepSeek-VL2-Tiny & 3.4B & 68.47 & 70.90 & 72.30 & 64.20 & 88.46 & 74.60 & 39.60 & 80.50 & -- & 42.90 & 81.00 & 88.90 & -- & 49.80 \\
MiMo-VL-7B-RL & 7B & 68.79 & 80.70 & -- & 72.68 & 83.33 & 82.80 & 63.80 & 82.90 & 37.80 & 26.40 & 91.70 & 95.70 & -- & 38.90 \\
Phi-4-MultiModal & 5.6B & 70.45 & 77.20 & 73.20 & 64.10 & 97.50 & 83.00 & 40.50 & 84.40 & 36.00 & 56.00 & 81.40 & 93.20 & -- & 58.90 \\
InternVL3.5-4B & 4B & 70.47 & 80.30 & -- & 66.30 & -- & 82.60 & 44.80 & 82.20 & 38.50 & 66.60 & 86.00 & 92.40 & -- & 65.00 \\
LLaVA-OneVision-7B & 7B & 70.91 & 85.13 & 76.54 & 66.93 & 95.93 & 81.41 & 61.71 & 72.50 & 38.25 & 51.93 & 80.84 & 87.39 & 62.04 & 61.20 \\
Qwen2.5-VL-7B & 7B & 72.71 & 82.20 & 77.00 & 68.50 & 87.70 & 84.30 & 51.90 & 88.80 & 38.80 & 58.00 & 87.30 & 95.70 & 60.88 & 64.10 \\
MiniCPM-V-4 & 4B & 73.62 & 79.70 & -- & 68.50 & -- & 82.90 & 50.80 & 89.40 & -- & 51.20 & 84.40 & 92.90 & -- & 62.80 \\
Kimi-VL-A3B-Instruct & A3B & 74.93 & 83.10 & 76.80 & 68.10 & 95.30 & 84.90 & 48.40 & 86.70 & -- & 57.00 & 87.70 & -- & -- & 61.30 \\
Qwen3-VL-8B & 8B & 75.59 & 88.66 & 78.71 & 69.54 & 94.34 & 83.79 & 72.42 & 89.00 & 41.25 & 56.83 & 86.24 & 95.64 & 61.89 & 64.40 \\
SAIL-VL-8B & 8B & 75.79 & 79.50 & 75.50 & 71.90 & 98.20 & 83.70 & 52.20 & 83.50 & -- & 48.20 & 84.60 & 92.20 & -- & 64.20 \\
Ovis2-4B & 4B & \underline{76.58} & 81.40 & 76.20 & 71.10 & 94.00 & 85.70 & 53.80 & 91.10 & -- & 49.00 & 84.20 & 94.00 & -- & 61.90 \\
\midrule
\multicolumn{16}{l}{\textit{Baseline and our model}} \\
Baseline & 4B & 74.12 & 87.59 & 78.74 & 71.37 & 92.36 & 81.76 & 69.39 & 88.09 & 39.45 & 52.34 & 84.84 & 94.72 & 61.98 & 60.93 \\
\textbf{CONVCUE (ours)} & \textbf{4B} & \textbf{78.82} & \textbf{92.57} & \textbf{82.49} & \textbf{75.03} & \textbf{95.39} & \textbf{84.97} & \textbf{76.08} & \textbf{89.41} & \textbf{43.72} & \textbf{60.62} & \textbf{87.08} & \textbf{96.20} & \textbf{73.09} & \textbf{68.00} \\
\bottomrule
\end{tabular}%
}
\end{table*}

\subsection{Main Results}
\label{sec:main_results}

Table~\ref{tab:vlm_comparison} and
Fig.~\ref{fig:capability_ablation}(a) show that \name{}-4B
improves over its original Qwen3-VL-4B backbone across all
13 evaluated benchmarks. The gains span GQA and HallusionBench, multimodal reasoning benchmarks such as MMMU and MMStar, and document and chart
understanding. This breadth shows that the benefits of the complete
augmentation framework are not confined to a single task
category, even when applied to an already capable VLM.

These improvements also make \name{}-4B competitive with
larger models: it outperforms Qwen3-VL-8B on all 13 benchmarks.
Its advantage is therefore consistent across the evaluated
tasks rather than driven by a few exceptional results.
While this comparison does not isolate architecture from
training differences, it shows that augmenting an existing
VLM can yield strong downstream performance without moving
to a larger language backbone.
Together, these results support our approach of combining
supplementary visual representations, learned fusion, and
joint adaptation to improve the complete model.

\begin{table*}[t]
    \centering
    \caption{Comparison with other baselines. Avg.\ denotes the unweighted mean over these nine benchmarks.
    Bold indicates the highest score in each column.}
    \label{tab:related_method_comparison}
    \setlength{\tabcolsep}{4pt}
    \renewcommand{\arraystretch}{1.10}
    \resizebox{\textwidth}{!}{%
    \begin{tabular}{l*{10}{c}}
        \toprule
        Method & Avg.
        & MMBench & SEED$^{\mathrm{I}}$ & RealWorldQA
        & SQA$^{\mathrm{I}}$ & AI2D & OCRBench
        & MMMU & ChartQA & DocVQA \\
        \midrule
        Cambrian-1-8B
        & 69.38
        & 75.90 & 74.70 & 64.20
        & 80.40 & 73.00 & 62.40
        & 42.70 & 73.30 & 77.80 \\

        LLaVA-UHD-v3-7B
        & 79.69
        & 81.30 & 77.20 & 70.30
        & \textbf{97.00} & 82.90 & 82.70
        & 50.20 & 82.80 & 92.80 \\

        Eagle-X4-8B-Plus
        & 72.42
        & 75.90 & 76.30 & 66.50
        & 84.30 & 76.10 & 62.60
        & 43.40 & 80.10 & 86.60 \\

        Florence-VL-8B
        & 71.34
        & 76.20 & 74.90 & 64.20
        & 85.90 & 74.20 & 63.40
        & 43.70 & 74.70 & 84.90 \\

        \midrule
        \name{}-4B (Ours)
        & \textbf{84.86}
        & \textbf{92.57} & \textbf{82.49} & \textbf{75.03}
        & 95.39 & \textbf{84.97} & \textbf{89.41}
        & \textbf{60.62} & \textbf{87.08} & \textbf{96.20} \\
        \bottomrule
    \end{tabular}%
    }
\end{table*}

To distinguish the gains of \name{} from those attributable
to additional training alone, we include matched Direct SFT
controls for Qwen3-VL-2B, Qwen3-VL-4B, and LLaVA-OneVision-7B.
Each control follows the same two-stage training procedure
as \name{}, using the same data at each stage, unfreezing
settings for the original model components, and training
hyperparameters, but without \name{}.
This comparison evaluates \name{} as an integrated
augmentation design. As shown in Fig.~\ref{fig:capability_ablation}(b),
Direct SFT provides only modest average gains on the two
Qwen3-VL backbones, whereas \name{} achieves larger improvements
and outperforms both the original models and the matched
training controls.
The difference is more pronounced on LLaVA-OneVision-7B:
Direct SFT slightly reduces its average score, while \name{}
substantially improves it.
Across all three backbones, the same data and training
procedure without \name{} do not reproduce the gains
of the complete framework.
These results support the effectiveness of the integrated
augmentation architecture together with joint adaptation,
with benefits extending across model sizes and families.
Complete per-benchmark comparisons are provided in
Appendix Table~\ref{tab:matched_sft_comparison}.

To evaluate \name{} against existing approaches to visual
representation enhancement, we compare it with Cambrian-1-8B,
Eagle-X4-8B-Plus, Florence-VL-8B, and LLaVA-UHD-v3-7B
on nine shared benchmarks.
As shown in Table~\ref{tab:related_method_comparison},
\name{}-4B achieves the highest average score and outperforms
all four methods on eight benchmarks, with ScienceQA being
the exception.
Its advantages cover both general visual understanding
and text, chart, and document understanding, rather than
depending on a few isolated results.
These comparisons show that retaining the original visual
pathway and adding a jointly adapted augmentation branch
can achieve competitive performance against existing visual
enhancement methods, even with a smaller language backbone.

\section{Analysis}
\label{sec:analysis}

\begin{figure}[t]
    \centering
    \includegraphics[width=\textwidth]{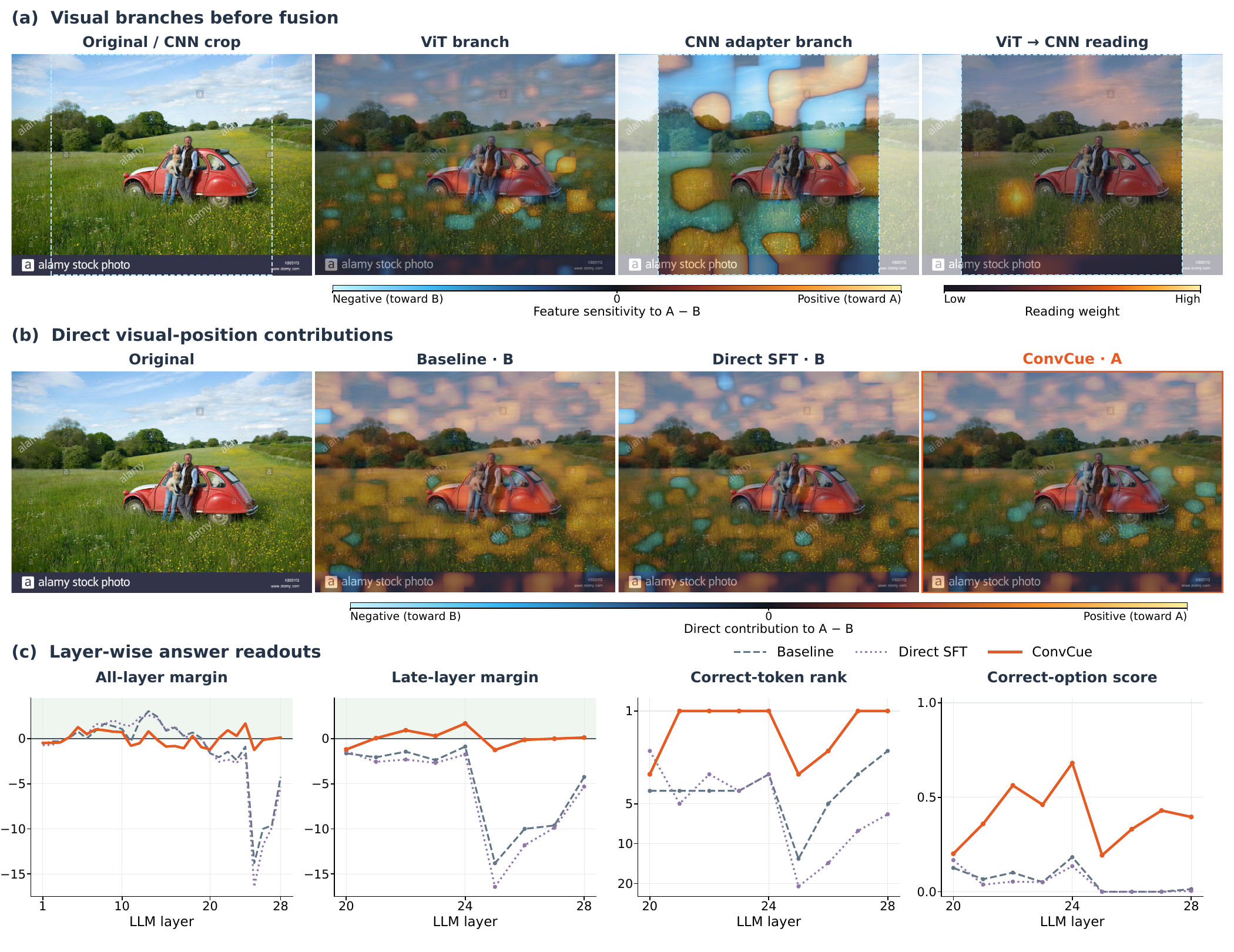}
    \caption{
        An example asking where the red car
        is located. The correct answer is A (centered), while B
        denotes the foreground left.
        \textbf{(a)} Original image with the CNN crop boundary,
        ViT and CNN feature sensitivities in \name{}, and
        cross-attention reading weights over CNN features.
        \textbf{(b)} Original image and direct visual-position
        contributions from tokens for the baseline,
        direct SFT, and \name{}.
        Warm/cool colors in signed maps indicate positive/negative
        values toward A relative to B; brighter reading weights
        indicate stronger attention.
        \textbf{(c)} Layer-wise readouts: correct-versus-best-incorrect
        logit margin, its late-layer detail, correct-token vocabulary
        rank, and correct-option softmax score.
    }
    \label{fig:seedbench_case}
\end{figure}

We first measure the size of the updates that \name{} adds
to the original visual features. We then examine visual
feature responses in \name{} and use a case study to compare
how answer preferences change across language-model layers
in the original model, the matched-training baseline,
and \name{}.

\subsection{Magnitude of Visual Representation Updates}
\label{sec:update_magnitude}

We measure how much the gated fusion module adds to the
original visual features. For image $i$, we compute
$r_i = \|g \odot A_i\|_F / \|X_i\|_F$, where $X_i$ denotes
the global-view visual tokens before fusion and $A_i$ is
the cross-attention output. Both are taken from the same
forward pass before the fusion LayerNorm.
On an ERQA subset, LLaVA-OneVision-7B equipped with \name{}
yields a mean ratio of 18.40\% and a median of 18.16\%,
with the 10th--90th percentiles ranging from 16.15\% to
20.51\%. The added residual is therefore substantial
relative to the original feature magnitude across the
analyzed samples.

\subsection{Case Study of Visual Responses and Answer Preferences}
\label{sec:answer_formation}

\noindent\textbf{Visual branches before fusion.}
Figure~\ref{fig:seedbench_case}(a) compares the two visual
branches in the trained \name{} model on a question about
the red car's location. The correct answer is A (centered),
while B denotes the foreground left.
We extract the ViT features and adapted CNN features before
fusion and compute their sensitivity to the answer-logit
difference $s = z_A - z_B$ at the first response position.
For the feature vector $F_i$ at spatial position $i$, we use
the signed gradient-times-activation score
$R_i = \sum_k F_{i,k}\,\partial s/\partial F_{i,k}$,
where $k$ indexes channels.
A positive score means that slightly scaling up this feature
vector would increase the preference for A over B; a negative
score means the opposite.
We map the scores to each branch's image grid using a shared
color scale, restricting the CNN map to its input crop. We also visualize where the fusion module reads CNN features.
For attention weight $\alpha_{hij}$ from ViT query $i$ to
CNN position $j$ in head $h$, we compute
$\bar{\alpha}_j = (HN)^{-1}
\sum_{h=1}^{H}\sum_{i=1}^{N}\alpha_{hij}$.
Brighter regions indicate larger average attention weights.
These weights depend on the image features, not the question,
and do not indicate support for either answer. The two branches show different spatial sensitivity patterns.
The fusion module also assigns relatively high attention
weights to surrounding regions, including the grass below
and to the left of the car, near positive responses in the
CNN sensitivity map.
In this example, feature retrieval therefore extends beyond
the target object to its surroundings.

\noindent\textbf{Direct visual-position contributions.}
We next examine how visual tokens contribute to answer
preference inside the language model.
Figure~\ref{fig:seedbench_case}(b) compares the original model,
the matched-training baseline, and \name{} on the same
image and question.
Before generating the first response token, we compute
$D_i = \sum_{\ell,h} a_{qi}^{(\ell,h)}
\langle \mathbf{u}_{A-B},
W_O^{(\ell,h)}v_i^{(\ell,h)} \rangle$,
where $q$ is the final prompt position,
$a_{qi}^{(\ell,h)}$ is its attention weight to visual token $i$,
and $v_i^{(\ell,h)}$ and $W_O^{(\ell,h)}$ are the value vector
and output projection for head $h$ in layer $\ell$.
The vector $\mathbf{u}_{A-B}$ is the difference between the
output-head vectors for A and B, adjusted by the final
RMSNorm scaling from the same forward pass.
We map scores for whole-image-view tokens back to the image
using a shared color scale: warm colors indicate direct
contributions toward A over B, and cool colors indicate
the opposite.
These scores account for attention weights and value content,
but exclude indirect effects through subsequent computation. Both comparison models show direct contributions toward
the correct option A in parts of the sky and grass,
yet ultimately select B.
Their errors therefore do not imply that visual tokens
provide no support for the correct answer; these contributions
account for only part of the final answer score.
\name{} shows a different distribution of positive and
negative contributions around the car and ultimately selects A,
without uniformly increasing support for A at every position.
These maps reveal how visual positions directly contribute
to the option-score difference, but do not by themselves
explain the final answer choice.
We therefore next examine how answer preferences change
across language-model layers.

\noindent\textbf{Layer-wise answer readouts.}
We next compare how answer preferences change across
language-model layers in the original model, the matched-training
baseline, and \name{} (Fig.~\ref{fig:seedbench_case}(c)).
After each decoder layer $\ell$, we extract the hidden state
$h_q^{(\ell)}$ at the final prompt position and compute
$z^{(\ell)} =
\operatorname{Head}(\operatorname{Norm}(h_q^{(\ell)}))$
using each model's own final normalization and output head.
These are readouts from intermediate representations,
not answers generated at each layer.
The final-layer readouts match the model's native output logits. We measure whether the correct option A leads the other
options using
$m^{(\ell)} = z_A^{(\ell)}
- \max_{b \in \mathcal{O}\setminus\{A\}} z_b^{(\ell)}$,
where $\mathcal{O}$ is the answer-option set.
A positive margin means that A ranks above every incorrect
option.
We also report A's token rank in the full vocabulary,
allowing ties, and its option-relative softmax score,
$p_A^{(\ell)} = \exp(z_A^{(\ell)})/
\sum_{b\in\mathcal{O}}\exp(z_b^{(\ell)})$.
This score measures preference among the options, not a
calibrated probability of correctness.
We focus on how preferences change within each model,
rather than comparing raw logit magnitudes across models. Both comparison models favor A in some intermediate layers,
but this preference does not persist to the final output.
Around layer 25, their margins fall and A's vocabulary
ranking worsens; both ultimately select B.
\name{} also shows a decline around this layer, but A's
ranking subsequently recovers, and A leads the other options
at the final layer.
Thus, the difference in this case is not whether a correct
answer can be read from an intermediate layer:
all three models show such a preference.
Rather, \name{} recovers the correct-answer preference
in later layers and outputs A, whereas the comparison
models ultimately favor B.

\section{Related Work}
\label{sec:related_work}

Prior work enhances VLM visual representations through image
processing, encoder combinations, and feature fusion.
LLaVA-UHD preserves high-resolution information through
image slicing and token compression \citep{guo2024llava},
while S$^2$ extracts features at multiple image scales
\citep{shi2024needlargervisionmodels}.
Cambrian-1 and Eagle study how features from different visual
encoders can be combined and connected to language models
\citep{tong2024cambrian, shi2025eagleexploringdesignspace}.
Florence-VL instead combines features from different depths
and task prompts of a generative vision model
\citep{chen2024florencevlenhancingvisionlanguagemodels}.
These studies explore different ways to obtain and integrate
visual representations. Our work follows this direction
by studying a supplementary convolutional branch together
with its fusion and adaptation to an existing VLM.

Convolutional representations have also been explored in VLMs.
ConvLLaVA replaces the ViT with a hierarchical ConvNeXt encoder
\citep{ge2024convllava}, while LLaVA-HR integrates high-resolution
CNN features into a low-resolution ViT pathway through
mixture-of-resolution adapters \citep{luo2025feast}.
\name{} instead augments the native visual representations of
an existing pretrained VLM while retaining its original encoder.
Its native visual tokens serve as queries to selectively retrieve
features from a frozen pretrained CNN through gated cross-attention.
This design incorporates supplementary convolutional information
without requiring one-to-one spatial correspondence or increasing
the number of visual tokens passed to the language model.

\section{Conclusion and Limitations}
\label{sec:conclusion}

We introduced \name{}, a visual representation augmentation
framework that integrates pretrained convolutional features into
existing VLMs. It retains the original visual encoding pathway
and uses a frozen CNN branch to provide supplementary
representations. A learnable adapter and gated cross-attention
allow native visual tokens to retrieve and incorporate these
features before passing through the original visual-to-language
mapping. Together with two-stage adaptation, this design enables
the model to use supplementary convolutional information without
replacing its original visual encoder or increasing the number
of projected visual tokens. Experiments on Qwen3-VL-2B, Qwen3-VL-4B, and
LLaVA-OneVision-7B across 13 benchmarks demonstrate higher
average performance than both the original models and matched
fine-tuning controls. On Qwen3-VL-4B, \name{} improves over
the original model on all 13 benchmarks, with gains spanning
visual question answering, document and chart understanding,
and multimodal reasoning. Pilot experiments inform the fusion
and visual-side adaptation choices, while further analyses
characterize representation updates, visual-position contributions,
and layer-wise answer preferences. These findings support
pretrained convolutional representations as a useful complement
to the native representations of modern VLMs and demonstrate
the effectiveness of incorporating them through learned
adaptation and fusion. Our experiments focus on image-based tasks and use a single
pretrained CNN backbone at a fixed input resolution. The
performance and efficiency trade-offs of other CNN architectures
and input resolutions remain to be evaluated. Extending the
framework to video also requires investigating how supplementary
features should be integrated across frames. Future work will
explore these settings to assess the broader applicability of
convolutional representation augmentation.

\bibliographystyle{iclr2027_conference}
\bibliography{iclr2027_conference}

\appendix
\clearpage

\newpage

\begin{center}
    \Large{\textbf{Technical Appendices}}
\end{center}

\vspace{1em}

\section{Experimental Settings}
\label{app:experimental_settings}

\noindent\textbf{Implementation details.}
We implement \name{} on Qwen3-VL-2B, Qwen3-VL-4B, and LLaVA-OneVision-7B, retaining the image processor and tokenizer associated with each backbone. The supplementary branch uses a pretrained ConvNeXt-Base and extracts its final-stage feature map. The CNN remains frozen throughout both training stages. CNN inputs are resized using bicubic interpolation,
center-cropped to $336\times336$, and normalized using the ImageNet mean and standard deviation. The fusion module uses eight attention heads, with channel-wise gate logits initialized to $-3$. The backbone operates in BF16, while the CNN branch,
adapter, and fusion module use FP32. Training is conducted on NVIDIA A100 GPUs with gradient checkpointing enabled. All the experiments are done with A100 GPUS.

\medskip
\noindent\textbf{Training data.}
The first stage uses image--text pairs from the LLaVA pretraining dataset for vision-language alignment. The second stage uses visual instruction examples from uhd \citep{guo2024llava} for supervised fine-tuning. The same stage-specific training data are used for \name{} and its matched Direct SFT control. 

\medskip
\noindent\textbf{Training configuration.}
In the first stage, we freeze the language model and train the original visual encoder, visual-to-language mapping modules, CNN adapter, and fusion module. In the second stage, we initialize from the first-stage checkpoint and additionally unfreeze the language model. We train for one epoch per stage with a per-GPU micro-batch size of 2 and eight gradient accumulation steps, giving an effective batch size of $16G$ for $G$ training GPUs. Both stages use AdamW with $(\beta_1,\beta_2)=(0.9,0.95)$, weight decay of $0.01$, and cosine learning-rate decay with a 3\% warmup ratio. The learning rates are $10^{-4}$ for the CNN adapter and fusion module, $10^{-6}$ for the original visual encoder, and $10^{-5}$ for the main Qwen visual merger or the LLaVA-OneVision projector. The Qwen DeepStack mergers use $5\times10^{-6}$. The language model uses $10^{-5}$ when unfrozen in the second stage.
The maximum training sequence length is 2,048, the gradient clipping norm is 1.0, and the random seed is 42.

\medskip
\noindent\textbf{Matched-training controls.}
For each backbone, Direct SFT follows the same two-stage procedure as \name{}, but without \name{}. The controls use the same data, training duration, batch configuration, and optimization settings. The original visual encoder and visual-to-language mapping modules are trained in both stages, while the language model is frozen in the first stage and unfrozen in the second.
Each control starts its second stage from its own first-stage checkpoint. Learning rates for the original model components are matched to those used in \name{}. The original-model baseline receives no additional training.

\medskip
\noindent\textbf{Inference and answer extraction.}
For models evaluated with our implementation, we use each backbone's native chat template and disable stochastic sampling.
The evaluation code limits generation to 8 new tokens for AI2D, SEEDBench, HallusionBench, ScienceQA, ERQA,
MMBench, and MMMU; 16 for RealWorldQA and GQA; 32 for OCRBench, ChartQA, and DocVQA; and 64 for MMStar. Multiple-choice prompts request an option letter, HallusionBench prompts request a yes/no answer, and text-answer tasks request a short answer. Responses are processed using task-specific answer extraction and normalization rules. The training implementation defaults to SDPA, while evaluation defaults to FlashAttention~2 with length-based batching.

\medskip
\noindent\textbf{Evaluation splits and scoring.}
The evaluation implementation loads the test splits for AI2D, SEEDBench, RealWorldQA, ScienceQA, OCRBench, ERQA, and ChartQA; the development split for MMBench; the validation splits for MMMU, DocVQA, and MMStar; the balanced test-dev split for GQA; and the image subset of HallusionBench. The implementation computes choice accuracy for multiple-choice tasks, question-level accuracy for HallusionBench, normalized answer matching for RealWorldQA and GQA, relaxed accuracy for ChartQA, and ANLS for DocVQA. It also outputs alternative matching scores for several text-answer tasks.

\medskip
\noindent\textbf{Score aggregation and published comparisons.}
All reported scores are expressed on a 0--100 scale. For the original models, Direct SFT controls, and \name{}, the overall average is the unweighted arithmetic mean across the 13 benchmarks. For Table~\ref{tab:related_method_comparison}, we separately compute the unweighted mean over the nine benchmarks shared by all listed methods. Results for Cambrian-1-8B, LLaVA-UHD-v3-7B, Eagle-X4-8B-Plus, and Florence-VL-8B are tested from the same split.

\section{Spatial QA Construction}
\label{app:spatial_qa_construction}

We construct Spatial QA from images and object instance annotations in the MS COCO 2017 training split for our pilot experiments \citep{lin2015microsoftcococommonobjects}. The task asks the model to identify the image region containing an object of a specified category. Questions and answers are generated automatically from category names and bounding boxes, without additional manual question--answer annotation.

\noindent\textbf{Position labels and question construction.}
For an image of width $W$ and height $H$ and an object bounding box $(x,y,w,h)$, we compute the normalized center coordinates as $u=(x+w/2)/W$ and $v=(y+h/2)/H$. We divide the image into a $3\times3$ grid using boundaries at $1/3$ and $2/3$ along each axis. We retain only objects whose centers fall within the four corner cells or the central cell, assigning one of five labels: \texttt{top-left}, \texttt{top-right}, \texttt{bottom-left}, \texttt{bottom-right}, and \texttt{center}. Objects with centers in the remaining four cells or exactly on a grid boundary are excluded. Labels depend only on the bounding-box center; the entire object need not lie within the assigned region. Each question follows the template ``Where is the [category] located in the image?'', with [category] replaced by the corresponding COCO category name.

\noindent\textbf{Sample filtering.}
To reduce ambiguity in referring to an object by its category, we exclude any image--category pair associated with multiple annotated instances. We also discard objects whose bounding-box area is less than $0.5\%$ of the image area and retain only samples with available image files. An image may yield multiple questions about different categories, but each retained image--category pair contributes exactly one question--answer example.

\noindent\textbf{Data splits and sampling.}
Using random seed 42, we shuffle the IDs of images containing valid examples and divide them into training, validation, and test sets in an $8:1:1$ ratio. We then collect the corresponding question--answer examples and perform sampling independently within each split. For training, we retain at most 8,000 examples per position label, keeping all examples for labels with fewer samples. For validation and testing, we downsample each position label to the count of the least frequent label within that split. The training set is therefore not strictly balanced, whereas the validation and test sets contain equal numbers of examples for all five labels.

The resulting training set contains 34,395 question--answer examples from 27,318 images. The validation set contains 3,290 examples from 2,695 images, with 658 examples per position label. The test set contains 3,175 examples from 2,629 images, with 635 examples per label. We verify that the final splits share no images. This dataset supports small-scale exploration of feature fusion and visual-side adaptation, with a task scope limited to coarse absolute object localization within an image.

\section{Ablation Study}
\label{app:design_ablations}

\subsection{Additional Details of Pilot Experiments}
\label{app:additional details of pilot}

\noindent\textbf{Experimental setup.}
We provide implementation details for the pilot experiments in Section~\ref{sec:pilot_experiments}. All experiments use LLaVA-1.5-7B and the same COCO-derived Spatial QA dataset. Dataset construction and splits are described in Appendix~\ref{app:spatial_qa_construction}. The language model and pretrained CNN remain frozen throughout these experiments. We report overall and per-position answer accuracy in
Fig.~\ref{fig:pilot_spatial_qa}.

\medskip
\noindent\textbf{Fusion configurations.}
We compare weighted fusion with gated cross-attention, training the original visual encoder, projector, CNN adapter, and fusion module in both settings. The data splits and training hyperparameters are kept the same.

For weighted fusion, a $1\times1$ convolution projects the CNN features into the ViT feature dimension. Adaptive average pooling then maps the feature grid to $24\times24$, matching the 576 visual tokens. After flattening, the adapter applies LayerNorm, a residual MLP, and a second LayerNorm. The aligned features are combined as
$\mathbf{z}_i =
\boldsymbol{\alpha}\odot\mathbf{v}_i +
(1-\boldsymbol{\alpha})\odot\mathbf{c}_i$,
where $\mathbf{v}_i$ and $\mathbf{c}_i$ are the ViT
and adapted CNN features at the corresponding grid
position, and
$\boldsymbol{\alpha}=\sigma(\mathbf{a})$.
The channel-wise mixing weights are shared across positions and images and initialized to $0.9$, initially assigning greater weight to the ViT features.

For cross-attention, the adapter retains the CNN feature grid without spatial resizing. The original visual tokens serve as queries, and the adapted CNN tokens serve as keys and values. The retrieved features are added through a channel-wise gated residual connection, followed by LayerNorm. This configuration allows each visual token to attend to all CNN positions without requiring equal token counts. The comparison therefore covers these two fusion configurations, including their respective spatial
alignment procedures.

\medskip
\noindent\textbf{Trainable-module configurations.}
With cross-attention fusion fixed, we compare three sets of trainable components: the CNN adapter and fusion module only; these added modules together with the original projector; and these components together with the original visual encoder. Thus, the ``adapter-only'' setting includes both the adapter and fusion module, while leaving all original VLM parameters frozen. The language model and CNN remain frozen even in the configuration with the largest trainable scope. These settings correspond to A, A+P, and A+P+E in Fig.~\ref{fig:pilot_spatial_qa}, respectively. The comparisons guide the fusion and visual-side adaptation choices explored in the subsequent framework.

\subsection{Generalization Across Model Backbones}
\label{app:backbone_generalization}

To examine whether the benefits of \name{} extend beyond our main backbone, Qwen3-VL-4B, we also apply it to Qwen3-VL-2B and LLaVA-OneVision-7B, covering different model sizes and architectures.
Across the 13 benchmarks, \name{} improves the average score from 69.65 to 72.64 on Qwen3-VL-2B, from 74.12 to 78.82 on Qwen3-VL-4B, and from 70.91 to 80.66 on LLaVA-OneVision-7B.

The improvements at both Qwen3-VL scales indicate that the benefits of convolutional feature augmentation are not restricted to a single model capacity.
The improvement on LLaVA-OneVision further supports its applicability across model families.
Although the magnitude of improvement varies across backbones, these results show that \name{} provides useful supplementary visual features while retaining each backbone's original visual encoding pathway, rather than offering benefits specific to Qwen3-VL-4B.

\subsection{Comparison with Direct Supervised Fine-Tuning}
\label{app:direct_sft_comparison}

To distinguish the contribution of the added architecture from the benefits of continued training, we include a fine-tuning control without convolutional augmentation, denoted as Direct SFT, for all three backbones.
This control uses the same data, two-stage training procedure, training hyperparameters, and unfreezing strategy for the original model components as \name{}.
In the first stage, both settings use LLaVA-pretrain data to update the visual encoder and projector or merger while keeping the language model frozen.
In the second stage, both use the same 529K training examples and additionally unfreeze the language model for joint fine-tuning.
The architectural difference is the inclusion of the CNN branch, CNN adapter, and fusion module in \name{}.
Thus, Direct SFT also undergoes both training stages rather than only second-stage fine-tuning from the original model.

Table~\ref{tab:matched_sft_comparison} reports the complete results.
Direct SFT achieves average scores of 69.81 and 75.00 on Qwen3-VL-2B and Qwen3-VL-4B, respectively, improving over their original baselines.
On LLaVA-OneVision-7B, however, its average score is 70.52, slightly below the original baseline.
In contrast, \name{} achieves 72.64, 78.82, and 80.66 on the three backbones, consistently outperforming the corresponding Direct SFT controls.
Under these matched training conditions, continued training alone does not reproduce the gains of \name{}, supporting the additional value of the convolutional augmentation architecture.

The advantage is not driven solely by a small number of benchmarks.
Compared with Direct SFT, \name{} obtains higher scores on 10 benchmarks and ties on one for Qwen3-VL-2B, improves on 12 benchmarks for Qwen3-VL-4B, and outperforms the control on all 13 benchmarks for LLaVA-OneVision-7B.
In particular, improvements on GQA, RealWorldQA, and HallusionBench occur across all three backbones, spanning compositional question answering, real-world scene understanding, and hallucination assessment.
Although gains are not universal on every benchmark, their breadth across tasks and backbones supports the effectiveness of convolutional feature augmentation beyond the benefits of additional supervised training.

\begin{table*}[t]
    \centering
    \caption{Comparison under matched two-stage training settings across three backbones. Baseline denotes the original model without additional training. Direct SFT follows the same training procedure as ConvCue but omits the convolutional augmentation architecture. Avg.\ is the unweighted mean over the 13 benchmarks. Bold indicates the highest score within each backbone, including ties.}
    \label{tab:matched_sft_comparison}
    \setlength{\tabcolsep}{4pt}
    \renewcommand{\arraystretch}{1.10}
    \resizebox{\textwidth}{!}{%
    \begin{tabular}{l*{9}{c}}
        \toprule
        & \multicolumn{3}{c}{Qwen3-VL-2B}
        & \multicolumn{3}{c}{Qwen3-VL-4B}
        & \multicolumn{3}{c}{LLaVA-OneVision-7B} \\
        \cmidrule(lr){2-4}
        \cmidrule(lr){5-7}
        \cmidrule(lr){8-10}
        Benchmark
        & Baseline & Direct SFT & ConvCue
        & Baseline & Direct SFT & ConvCue
        & Baseline & Direct SFT & ConvCue \\
        \midrule
        AI2D
        & 76.02 & 76.26 & \textbf{78.59}
        & 81.76 & 81.90 & \textbf{84.97}
        & 81.41 & 80.76 & \textbf{90.28} \\
        SEEDBench
        & 75.90 & 76.02 & \textbf{79.94}
        & 78.74 & 77.99 & \textbf{82.49}
        & 76.54 & 76.56 & \textbf{84.94} \\
        HallusionBench
        & 63.07 & 62.88 & \textbf{69.93}
        & 69.39 & 69.62 & \textbf{76.08}
        & 61.71 & 63.83 & \textbf{72.24} \\
        RealWorldQA
        & 64.71 & 63.01 & \textbf{67.19}
        & 71.37 & 72.42 & \textbf{75.03}
        & 66.93 & 59.87 & \textbf{79.08} \\
        ScienceQA
        & 87.66 & 88.70 & \textbf{89.04}
        & 92.36 & \textbf{95.74} & 95.39
        & 95.93 & 96.92 & \textbf{97.97} \\
        OCRBench
        & 85.13 & \textbf{85.30} & 83.80
        & 88.09 & 87.50 & \textbf{89.41}
        & 72.50 & 75.60 & \textbf{88.20} \\
        ERQA
        & 35.93 & \textbf{39.75} & \textbf{39.75}
        & 39.45 & 40.75 & \textbf{43.72}
        & 38.25 & 40.00 & \textbf{50.25} \\
        MMBench
        & 83.06 & 83.09 & \textbf{87.11}
        & 87.59 & 87.82 & \textbf{92.57}
        & 85.13 & 85.26 & \textbf{90.48} \\
        MMMU
        & 47.76 & 46.46 & \textbf{48.76}
        & 52.34 & 56.77 & \textbf{60.62}
        & 51.93 & 49.06 & \textbf{65.34} \\
        ChartQA
        & 80.32 & 80.44 & \textbf{81.08}
        & 84.84 & 85.28 & \textbf{87.08}
        & 80.84 & 81.80 & \textbf{88.80} \\
        DocVQA
        & 92.47 & \textbf{93.35} & 93.25
        & 94.72 & 95.33 & \textbf{96.20}
        & 87.39 & 89.27 & \textbf{95.44} \\
        GQA
        & 58.94 & 58.53 & \textbf{66.93}
        & 61.98 & 61.40 & \textbf{73.09}
        & 62.04 & 58.71 & \textbf{74.85} \\
        MMStar
        & 54.53 & 53.73 & \textbf{58.93}
        & 60.93 & 62.53 & \textbf{68.00}
        & 61.20 & 59.07 & \textbf{70.67} \\
        \midrule
        Avg.
        & 69.65 & 69.81 & \textbf{72.64}
        & 74.12 & 75.00 & \textbf{78.82}
        & 70.91 & 70.52 & \textbf{80.66} \\
        \bottomrule
    \end{tabular}%
    }
\end{table*}

\begin{table}[!htbp]
    \centering
    \small
    \setlength{\tabcolsep}{6pt}
    \renewcommand{\arraystretch}{1.10}
    \caption{Inference efficiency with synthetic inputs.
    Latency is the median time to first token with inputs
    already on the GPU. Memory is the maximum peak allocated
    GPU memory across repeated measurements. Input length
    includes both visual and text tokens.}
    \label{tab:inference_efficiency}
    \begin{tabular}{llrrrr}
        \toprule
        & & \multicolumn{2}{c}{Latency (ms)}
        & \multicolumn{2}{c}{Peak memory (GiB)} \\
        \cmidrule(lr){3-4}
        \cmidrule(lr){5-6}
        Model & Input tokens
        & Baseline & \name{}
        & Baseline & \name{} \\
        \midrule
        Qwen3-VL-2B & 1024
        & 83.98 & 91.34 & 5.53 & 5.95 \\
        & 2048
        & 104.52 & 112.84 & 5.54 & 5.95 \\
        & 4096
        & 194.41 & 202.24 & 7.09 & 7.50 \\
        & 8192
        & 543.02 & 551.61 & 15.22 & 15.63 \\
        \midrule
        Qwen3-VL-4B & 1024
        & 101.78 & 109.46 & 9.89 & 10.31 \\
        & 2048
        & 160.54 & 168.36 & 10.01 & 10.42 \\
        & 4096
        & 386.87 & 394.74 & 14.13 & 14.55 \\
        & 8192
        & 1267.09 & 1276.66 & 29.95 & 30.36 \\
        \bottomrule
    \end{tabular}
\end{table}

\section{Inference Efficiency}
\label{app:inference_efficiency}

We evaluate the inference overhead of \name{} on Qwen3-VL-2B and Qwen3-VL-4B using a single NVIDIA A100 GPU. All measurements use eager attention, a batch size of one, and greedy generation of one token. We use a fixed synthetic image of $896 \times 896$ pixels, corresponding to 784 visual tokens after merging, and vary the total input length by adding synthetic text tokens. The CNN branch processes the same image at $336 \times 336$ resolution. The Qwen backbone uses BF16, while the CNN branch, adapter, and fusion module retain FP32, matching our evaluation implementation.

We measure time to first token with the input tensors already on the GPU, including visual encoding, feature fusion where applicable, language-model prefill, and token selection. Timing uses wall-clock measurements with CUDA synchronization and excludes preprocessing, data transfer, and model loading. After 10 warm-up generation calls, we collect 30 measurements per configuration and report the median latency. We reset the peak-memory statistics before each measurement and report the maximum peak allocated GPU memory across the 30 measurements, including model parameters and inference tensors.

As shown in Table~\ref{tab:inference_efficiency}, \name{} introduces a modest inference overhead under these settings. The additional latency ranges from 7.4 to 9.6 ms, while peak allocated memory increases by approximately 0.41 GiB across both model sizes and all input lengths. With image resolution held fixed, the added latency remains approximately constant as the text context grows, reducing its relative overhead at longer input lengths. These results show that the convolutional feature enhancement can be incorporated with a small additional memory footprint and limited first-token latency overhead in the evaluated setting.

\end{document}